\documentclass[pdflatex,sn-mathphys-num]{sn-jnl}

\usepackage{graphicx}%
\usepackage{multirow}%
\usepackage{amsmath,amssymb,amsfonts}%
\usepackage{amsthm}%
\usepackage[title]{appendix}%
\usepackage{xcolor}%
\usepackage{textcomp}%
\usepackage{manyfoot}%
\usepackage{booktabs}%
\usepackage{algorithm}%
\usepackage{algorithmicx}%
\usepackage{algpseudocode}%
\usepackage{listings}%
\usepackage{booktabs}
\usepackage{tabularx}
\usepackage{tcolorbox}
\usepackage{xurl}  

\theoremstyle{thmstyleone}

\theoremstyle{thmstyletwo}%

\theoremstyle{thmstylethree}%

\makeatletter
\newcommand{\blfootnote}[1]{%
  \begingroup
  \renewcommand{\thefootnote}{}%
  \def\Hy@footnote@currentHref{Hfootnote.blank}%
  \footnotetext{#1}%
  \endgroup
}
\makeatother

\begin{document}

\title[Evaluating VLM Coherence in Mammography]{Evaluating Cross-Signal Coherence in Mammography Vision-Language Models}

\author[1,2]{\fnm{Rebeca} \sur{Nam}}\email{rebeca.nam@emory.edu}

\author[2]{\fnm{Chase M.} \sur{Fensore}}\email{chase.fensore@emory.edu}
\author[3]{\fnm{Christian A.} \sur{Fauria-Robinson}}\email{christian.allison.fauria-robinson@emory.edu}
\author[3]{\fnm{Judy} \sur{Gichoya}}\email{judywawira@emory.edu}
\author[2]{\fnm{Joyce C.} \sur{Ho}}\email{joyce.c.ho@emory.edu}

\affil[1]{\orgname{Yonsei University}, \orgaddress{\street{50 Yonsei-ro}, \city{Seodaemun-gu}, \postcode{03722}, \state{Seoul}, \country{Republic of Korea}}}

\affil[2]{\orgdiv{Department of Computer Science}, \orgname{Emory University}, \orgaddress{\street{400 Dowman Drive}, \city{Atlanta}, \postcode{30322}, \state{GA}, \country{USA}}}

\affil[3]{\orgdiv{Department of Radiology and Imaging Sciences}, \orgname{Emory University}, \orgaddress{\street{100 Woodruff Circle}, \city{Atlanta}, \postcode{30322}, \state{GA}, \country{USA}}}

\abstract{As vision-language models (VLMs) move toward interpreting medical images directly, their outputs increasingly include not just a label but a stated confidence and a written justification. Yet the field still predominantly evaluates VLMs using just accuracy. We evaluate the coherence between answer, confidence, and explanation across four VLMs on breast density and BI-RADS categorization, using two mammography datasets. We perform controlled perturbations of the question, multiple choice option orders, and the images and assess changes in the three outputs. Our results suggest confidence does not decrease even as the mammography image is changed. Shuffling the options also changes most VLMs' answers. Notably, there are cases where a model's own stated reasoning contradicts its final answer, a distinction accuracy alone cannot capture but one that governs whether an AI-generated answer can be trusted.} 


\blfootnote{Code availability: \url{https://anonymous.4open.science/r/mammo-coherence-BB66}}

\maketitle

\section{Introduction}\label{sec:intro}

An aging population and rising chronic disease burden have placed increasing imaging demand on the U.S. radiology workforce, lengthening patient wait times and contributing to radiologist burnout \citep{jing2025ai}.
Breast cancer screening offers a case study for when this workforce strain has clinical consequence. 
The 2024 U.S. Preventive Services Task Force recommendation statement finds biennial mammography screening from ages 40–74 carries a moderate but real mortality benefit, with modeling estimating a 30\% relative reduction in breast-cancer mortality compared to no screening \citep{us2024screening}.  Thus any degradation in screening capacity or turnaround may negatively impact patient outcomes.

Most mammography AI today assists a human reader rather than replacing one, typically serving as a second reader in the standard double-reading workflow.
However, replacing both readers is a fundamentally different demand than replacing one. A second reader can assume a first read already exists and need only add a risk score or detection, whereas a system replacing both must produce an assessment \emph{with reasoning} a clinician can audit. This is a different capability than the AI models evaluated in the second-reader studies \citep{elhakim2024ai, warren2026impact}, which were typically trained to output a risk score or bounding box rather than a categorical clinical judgment with attached reasoning. This shift to autonomous roles has started to happen with at least one AI system receiving certification for autonomous mammogram triage, bypassing radiologist review for low-risk cases \citep{VaraReceivesWorldFirst}.

Vision-language models (VLMs), or models that take images and text as input to reason across both modalities and typically produce text as output, are a natural candidate for this automation task. When framed as a visual question answering (VQA) problem, a VLM can generate a category assessment (e.g., a density score), a confidence estimate, and a natural-language rationale over the screening mammography images, organically mirroring how a radiologist conducts a full read of a mammogram. VLM usage for chest x-rays has been studied heavily, with several VQA benchmarks and robustness analyses \citep{CheXVQA,MIMIC-CXR-VQA,pal2025rexvqa,fallahpour25a}, but these works largely score only the generated answer, and high performance on such a benchmark is not necessarily sufficient to establish VLM understanding or generalizability. Limited work assesses whether a chest X-ray VLM's \emph{stated reasoning} can be trusted rather than just its answer \citep{VisualGrounding} 
or whether existing explanation methods can identify the image regions that actually drove the model's prediction \citep{xiong2026rethinking}.

Each of those evaluations, however, still treats one signal at a time.
Separate strands of general-domain VQA research show that each signal is individually unreliable: answers are sensitive to superficial reformatting \citep{OrderSensitivity,Zheng2024_RobustMCQ}, models routinely hallucinate a choice rather than abstain \citep{Tong2025_HumbleBench, Demirci2026_NOTA}, and confidence and reasoning are each known to be poorly calibrated or unfaithful \citep{Tian2023,Kadavath2022, Matchmaker,Turpin2023,Lanham2023}.
What remains unseen is the extent to which the output signals (i.e., confidence-derived answer, confidence, and reasoning) \emph{agree with each other},
and whether all three are grounded in the visual evidence, when elicited jointly from the same response. This evidence is necessary for a radiologist to trust or audit an AI-generated read.

We study this question in mammography rather than chest
radiography, since its interpretation tasks offer several properties well suited to testing coherence. Its two standard interpretation tasks, (i) breast density with American College of Radiology (ACR) categories A–D, and (ii) BI-RADS assessment with categories 0–6, are ordinal and multi-category.  Thus a model's errors, confidence, and reasoning can be graded rather than simply right or wrong. Both label sets are standardized clinical lexicons defined by the ACR and follow a standard imaging protocol so behavior under perturbation can be cleanly attributed. Breast density is also subject to federal reporting and patient-notification requirements in the U.S., and BI-RADS categories directly inform clinical action, from routine recall to biopsy. Here, a model's confidence and reasoning shape what a patient is told and what happens next.

We test this using two subsets, RSNA and EMBED, from MammoVQA
\citep{MammoVQA}, the largest VQA benchmark for mammography. Both provide breast density and BI-RADS labels over a large, real screening population. We evaluate four open VLMs. MedGemma-4B and Lingshu-7B are medically trained, while Qwen2-VL-7B and InternVL3-8B are general-purpose. Rather than scoring a single response, we jointly elicit the answer, a per-option confidence score, and a natural-language rationale from the same output, then perturb every question. We replace, blank, or remove the mammogram to probe visual grounding, and reword the question or reorder the options to
probe elicitation robustness. A grounded model should grow less confident as the image degrades and reason consistently with the final assessment. Instead, across both datasets and all four models, we find the opposite. Reasoning is sometimes correct where the scored answer is wrong, confidence fails to track the evidence and at times moves the wrong way, models conflate appropriate abstention with overconfident guessing, and answers flip with the question's format rather than its content. None of these are visible from accuracy alone. 
\section{Related Work}

\textbf{Evaluating VLMs beyond accuracy.}
A large body of work probes VLMs beyond a single accuracy score. Reordering multiple-choice options or rephrasing prompts can flip answers and shift reported confidence \citep{OrderSensitivity, Zheng2024_RobustMCQ}. Perturbation-based protocols test visual dependence by corrupting, masking, or substituting images \citep{Hallucination, VisualGrounding}. Similarly, blind baselines and negation probes expose failures in medical VQA \citep{yanWorseRandomEmbarrassingly2024}. Adversarial ``None of the Above'' (NOTA) protocols reveal that models often hallucinate answers when no valid option exists, in both general multimodal domains \citep{Tong2025_HumbleBench} and medical diagnostics \citep{Demirci2026_NOTA}. These methods primarily assess answer correctness or visual dependence, whereas we examine how answers, confidence, and reasoning jointly respond to visual perturbations.

\textbf{Output coherence and confidence calibration.}
Faithfulness methods assess whether explanations reflect the model's internal decision process. Siegel et al.~\citep{siegelProbabilitiesAlsoMatter2024} use counterfactual edits to measure changes in predicted probabilities and whether explanations acknowledge those edits. Parcalabescu and Frank \citep{parcalabescuMeasuringFaithfulnessSelfconsistency2024} examine explanation faithfulness and self-consistency, while interventional tests probe chain-of-thought faithfulness by corrupting reasoning traces \citep{lanhamMeasuringFaithfulnessChainofThought}. These methods address internal faithfulness, whereas we assess whether a rationale agrees with its selected answer and the imaging evidence. Confidence calibration is likewise studied through aggregate calibration error \citep{duConfidenceCalibrationMultimodal2026} and methods that elicit fidelity across sampled responses \citep{zhangCalibratingConfidenceLarge2024}. In contrast, we examine whether confidence responds to visual evidence and agrees with the answer and rationale within a single response. Aggregate calibration and cross-sample consistency do not establish such within-response coherence.

\textbf{Medical VLM evaluation.}
VLMs for chest radiography are evaluated predominantly using report-derived accuracy metrics, including exact- or close-match accuracy on VQA pairs \citep{CheXVQA, MIMIC-CXR-VQA} and overlap between generated and reference reports \citep{CheXVQA}. Mammography-specific VLMs follow a similar pattern. MammoVLM \citep{cao2025mammovlm} reports performance approaching junior-radiologist-level performance on a private dataset, based on radiologist ratings of correctness, rationality, helpfulness, professionalism, and completeness. A separate reinforcement learning-based VLM \citep{li2026mammovqa} is trained on EMBED data using a domain-specific reward but evaluates performance primarily through answer correctness. Neither model's weights are publicly released.
\section{Methodology}\label{sec:method}

\begin{figure}[t]
\centering
\includegraphics[width=1\linewidth]{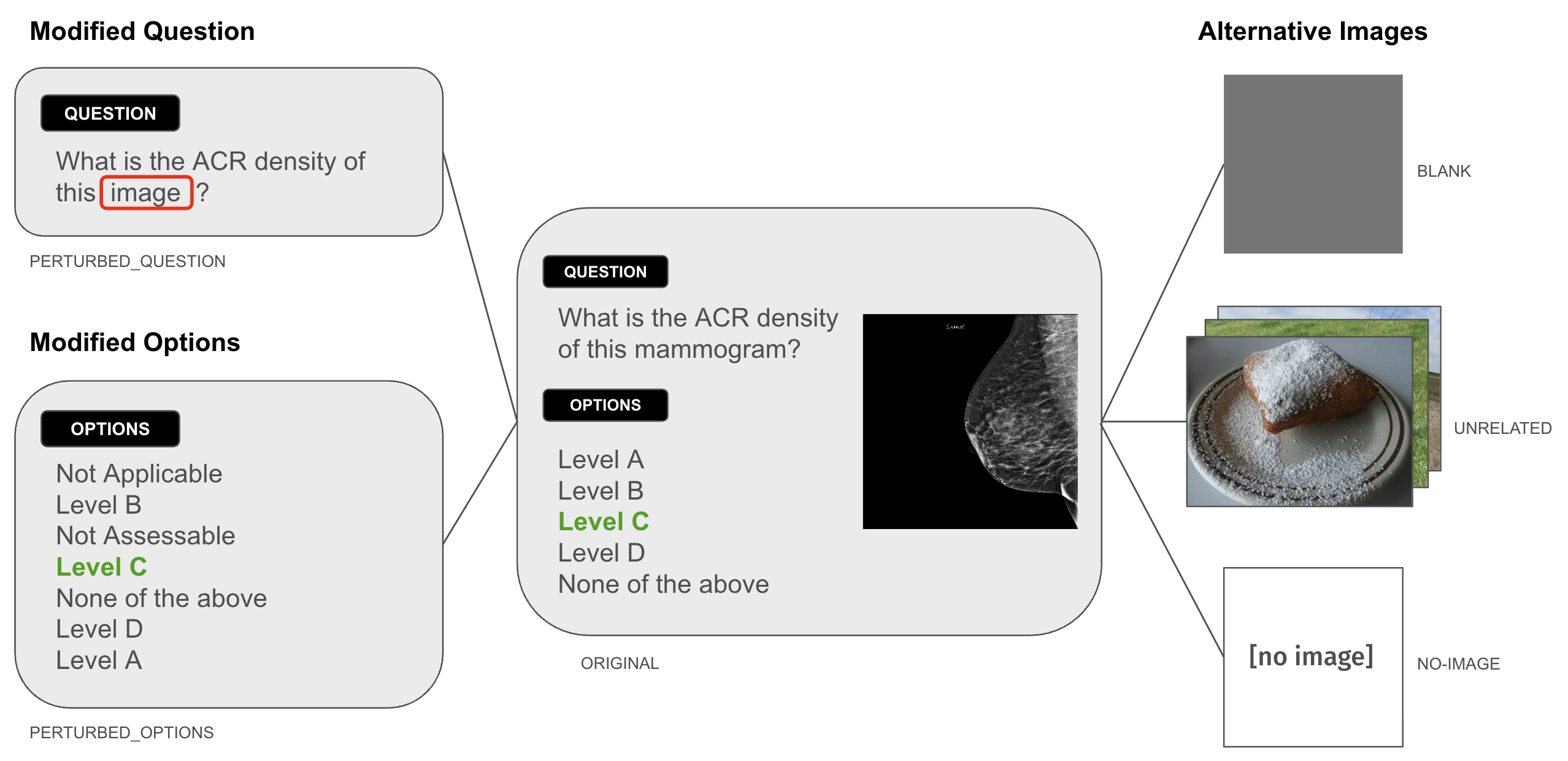}
\vspace{-5pt}
\caption{Our proposed evaluation methodology with six conditions: the original image and five perturbations.}
\label{fig:overview}
\vspace{-5pt}
\end{figure}

\subsection{Overview}
General VLM research probes accuracy, reasoning, and confidence individually with limited work on how these signals interrelate. There is limited work detecting when a model is confidently wrong, right in its explanation yet wrong in its answer, or unchanged by the removal of the image. We address this gap directly. For each visual question, a VLM produces three signals a clinician might rely on: the answer it selects, the confidence it reports, and the reasoning it gives. We jointly elicit all three from a single response and test whether they agree with each other and with the visual evidence, using mammography as a testbed, under five controlled perturbations, shown in Figure \ref{fig:overview}, known to affect model responses. We also provide the code to reproduce our results.

\subsection{Dataset and Tasks}
\label{sec:data}
We use the MammoVQA benchmark~\citep{MammoVQA}, focusing on its two largest cohorts, the RSNA screening mammography dataset~\citep{rsna2023screening} and EMBED~\citep{jeong2023emory-996}. We study two interpretation tasks: breast density (ACR categories A–D) and BI-RADS assessment (categories 0–6), both of which in theory require reading the image rather than exploiting coarse spatial cues. The RSNA screening cohort contains $N{=}5{,}106$ BI-RADS assessment and $N{=}5{,}821$ breast density / ACR items while EMBED contains $N{=}14{,}303$ BI-RADS assessment items. Each item is a single mammographic image (RSNA) or multiple exam images (EMBED) paired with a templated question, a fixed option set, and a benchmark label $y$ (detailed statistics in Appendix \ref{app:data}).

\subsection{Signal elicitation}
\label{sec:elicitation}
We elicit the answer, confidence, and reasoning jointly using a structured multiple-choice prompt with per-option confidence scores following the Matchmaker formulation~\citep{Matchmaker}.~\citep{Matchmaker}.
The original MammoVQA benchmark poses each question as a closed multiple-choice question (MCQ) with no option for rejecting the offered choices. Since we will perturb the original image, none of the original options are correct, so the existing options would force the model into an arbitrary wrong answer and provide no mechanism for distinguishing a grounded refusal from a guess. We therefore augment each question with an explicit \textbf{NOTA} (``\emph{None of the above}'') option, representing the claim
that no listed answer is correct. 

\begin{figure}[t]
\centering
\begin{tcolorbox}[colback=gray!5, colframe=gray!40, boxrule=0.5pt, arc=2pt]
\small
\textit{1) State your single best answer as the option letter.\\
2) For EACH option, give a confidence score from 0 to 100 for how likely THAT option is the correct answer (score each option independently; they need not sum to 100).\\
3) Give a brief one-sentence explanation of your reasoning.}
\end{tcolorbox}
\caption{Prompt used to jointly elicit an explicit answer, per-option confidence scores, and reasoning in a single response.}
\label{fig:prompt}
\end{figure}

Each question is thus presented as an MCQ over its $K$ options plus NOTA (Figure~\ref{fig:prompt}). From the response we recover three signals: the \textbf{answer} $\hat{y}$, the option the model explicitly states; the \textbf{confidence} $c = s_{\hat{y}}$, the score it assigns to that answer, from
the per-option scores $\mathbf{s}\in[0,100]^K$; and the \textbf{reasoning} $\hat{y}_e$, the answer named in the free-text explanation. The independent per-option scores do not force a normalized distribution, allowing the model to assign \emph{all} options low scores when no option is well supported by the evidence.

\subsection{Perturbation conditions}
\label{sec:conditions}
Each question is evaluated under 6 conditions: an \textsc{original}
baseline using the unmodified question and mammogram, and 5 conditions
that perturb the \emph{visual evidence} or the \emph{elicitation} while holding the question otherwise fixed. The $K$ options plus NOTA are presented under every condition. The 3 visual perturbations probe grounding: a VLM that reads the image should change its behavior when the image is made uninformative or removed. Under the \textsc{unrelated} condition, the mammogram is replaced with an unrelated natural photograph. In the \textsc{blank} scenario, it is replaced with a blank, grayscale image. For the \textsc{no-image} perturbation, the image is omitted entirely and the model answers from the question text alone, probing the model's language prior in the absence of any visual input.
The two elicitation perturbations hold the mammogram  fixed and instead alter how the question and answer are posed, isolating a format effect independent
of the evidence. Under the \textsc{perturbed-question} condition, any domain-specific terminology is neutralized (e.g., ``mammogram'' is replaced with
``image'') to probe sensitivity to wording. In \textsc{perturbed-options} scenario, the answer options are shuffled and augmented with paraphrase filler distractors (e.g., ``Not Assessable,'' ``Not Applicable'') alongside NOTA, probing sensitivity to option order and format.

\textbf{NOTA correctness.} In the three conditions with no valid
mammogram (\textsc{unrelated}, \textsc{blank}, \textsc{no-image}), the
appropriate answer is to select NOTA. We therefore define a second target in which the correct answer is NOTA, and report performance under
both the \emph{original label} (did the model recover the benchmark answer) and this \emph{NOTA-aware label} (did the model correctly decline). The former measures classification while the latter appropriate refusal.

\subsection{Coherence metrics}
\label{sec:metrics}
Let $\hat{y}$, $c$, and $\hat{y}_e$ denote the answer, confidence, and reasoning-answer respectively for a
given item and condition, and let $y$ be the benchmark label. The reasoning-answer $\hat{y}_e$ is extracted from the explanation with a rule-based matcher.

\textbf{Reasoning Accuracy.} We combine a label matcher with a definition-based semantic matcher. (i) The label matcher identifies explicit category mentions by locating task cues (``BI-RADS'' for assessment; ``level'', ``category'', ``ACR'', or ``density'' for density) and extracting the nearby category token (digit 0--6 for BI-RADS; letter A--D for density), using the last match when several occur. (ii) The semantic matcher maps descriptions of findings to categories using official definitions. For example, \ ``no abnormality / negative'' $\rightarrow$ BI-RADS 1 and ``almost entirely fatty / scattered fibroglandular / heterogeneously dense / extremely dense'' $\rightarrow$ A/B/C/D, while also identifying ``none of the above'' conclusions.

To validate the matcher we drew a random sample of $120$ explanations
(15 per model per task, stratified across the four models and both tasks) and manually annotated which category the explanation supports or ``none''/``unparsable''. The annotator was blind to the matcher output and the benchmark label. Agreement between the matcher and the human labels was substantial: Cohen's $\kappa = 0.78$ ($80.8\%$ exact agreement), and comparable on both tasks (BI-RADS $\kappa = 0.79$, $83.3\%$; density $\kappa = 0.73$, $78.3\%$). Unparseable explanations are excluded from the reasoning-accuracy denominator (see Appendix \ref{app:parser_examples} for examples of unparsable instances).

\textbf{Signal agreement.} We report how often the answer disagrees with the reasoning ($\hat{y} \neq \hat{y}_e$) and with the highest-scored option ($\hat{y} \neq \arg\max_k s_k$), and how often all three agree. A coherent model
should state the option it scores highest and describe that same option in its explanation.

\textbf{Confidence--evidence alignment.} For a given item and condition, let $c_y = s_y$ denote the score placed on the \emph{benchmark-correct} option $y$, and $c_{\hat{y}} = s_{\hat{y}} = c$ the score placed on the model's \emph{chosen} option. We track both, averaged per model and condition, across the evidence and elicitation perturbations. A grounded, well-calibrated model should have $c_y$ increase as informative visual evidence is restored, and $c_{\hat{y}}$ should track $c_y$ closely. We report the gap $c_{\hat{y}} - c_y$: a positive gap measures overconfidence in an incorrect selection, while a negative gap means the model scored the correct option higher than the answer it chose, which is itself an incoherence between answer and confidence.

\textbf{NOTA sensitivity.} We report two summary quantities per model and task. The \emph{real$\rightarrow$visual difference} is the NOTA-selection rate on the three visual perturbations (\textsc{unrelated}, \textsc{blank},
\textsc{no-image}), averaged, minus the rate on the real image. A positive value indicates the model abstains more often when evidence is genuinely absent. $\Delta_{\mathrm{NOTA}} = \mathrm{Acc}_{\mathrm{NOTA}} - \mathrm{Acc}_{\mathrm{base}}$ is the NOTA-aware accuracy minus the original-label accuracy. A large $\Delta_{\mathrm{NOTA}}$ paired with a small real$\rightarrow$visual difference
indicates blanket refusal rather than genuine evidence-sensitivity.

\textbf{Reasoning--answer accuracy gap.} On items with a parseable explanation, we compare reasoning accuracy ($\hat{y}_e = y$) to answer accuracy ($\hat{y} = y$) on the same items. A positive gap means the explanation identifies the correct label more often than the stated answer does; a negative gap means the reverse.

\textbf{Coherence taxonomy.} Each item with a parseable explanation is classified by whether $\hat{y}$ and $\hat{y}_e$ each equal $y$, giving four buckets:
\textbf{A} ($\hat{y}{=}y$, $\hat{y}_e{=}y$), coherent and correct; \textbf{B} ($\hat{y}{=}y$, $\hat{y}_e{\neq}y$), a correct answer with unsupported reasoning; \textbf{C} ($\hat{y}{\neq}y$, $\hat{y}_e{=}y$), a wrong answer with
correct reasoning; and \textbf{D} ($\hat{y}{\neq}y$, $\hat{y}_e{\neq}y$), both wrong. Restricting to confident errors ($c \ge 90$), the \emph{recoverable rate}
is the confident subset of \textbf{C} over the confident subsets of \textbf{C} and \textbf{D} combined: the share of a model's confident errors whose own reasoning already contains the right answer.

\subsection{Models}
\label{sec:models}
We evaluate four open VLMs spanning medical and general-purpose training: MedGemma-4B and Lingshu-7B (medical), and Qwen2-VL-7B and InternVL3-8B (general). These models were selected for three reasons: (1) open weights enable reproducibility without costly API access, excluding closed-source models (e.g., GPT-4V and Gemini Ultra) and mammography-specific systems \citep{cao2025mammovlm,li2026mammovqa} whose weights are unavailable; (2) the selection enables comparison of medical versus general-purpose training; and (3) the 4--8B parameter range, using unquantized models, limits differences attributable to model scale and quantization. All models receive an identical instruction; only the model-specific chat template and image-token convention differ (checkpoint and inference settings in Appendix \ref{app:model-settings}). Decoding is greedy for reproducibility.
\section{Results}\label{sec:results}

\begin{table}[t]
\centering
\scriptsize
\setlength{\tabcolsep}{2pt}
\renewcommand{\arraystretch}{0.9}
\caption{Per-task output accuracy and ``None of the above'' selection rate, with the majority-class prior.
The answer is the model's explicitly stated single-best choice.}
\label{tab:baseline}
\begin{tabular}{lrr}
\toprule
Model & Output acc.\ \% & NOTA sel.\ \% \\
\midrule
\multicolumn{3}{l}{\textit{RSNA -- BI-RADS (6 true classes; majority 59.4\%)}} \\
MedGemma-4B  & \textbf{42.4} & 0.0  \\
Lingshu-7B   & \underline{12.0} & 0.0  \\
Qwen2-VL-7B  & 2.9  & 62.8 \\
InternVL3-8B & 7.7  & 13.7 \\
\midrule
\multicolumn{3}{l}{\textit{RSNA -- Density (ACR) (4 true classes; majority 42.6\%)}} \\
MedGemma-4B  & 33.9 & 0.0  \\
Lingshu-7B   & \textbf{49.3} & 0.0  \\
Qwen2-VL-7B  & 24.8 & 33.2 \\
InternVL3-8B & \underline{42.0} & 0.0  \\
\midrule
\multicolumn{3}{l}{\textit{EMBED -- BI-RADS (6 true classes; majority 62.1\%)}} \\
MedGemma-4B  & \textbf{36.7} & 0.0  \\
Lingshu-7B   & 10.3 & 0.0  \\
Qwen2-VL-7B  & \underline{15.2} & 0.7  \\
InternVL3-8B & 5.1  & 0.1  \\
\bottomrule
\end{tabular}
\vspace{4pt}
\parbox{\linewidth}{\footnotesize
\textbf{Output acc.}: the explicit answer $\hat{y}$ equals ground truth $y$.
\textbf{NOTA sel.\ \%}: rate at which the model explicitly selects ``None of the above''.}
\end{table}

Table~\ref{tab:baseline} provides the overall picture for each task. Under direct elicitation, no model exceeds the majority-class prior on BI-RADS, while on density several approach or exceed it (e.g.\ Lingshu at $49.3\%$ vs.\ the $42.6\%$ prior). Explicit abstention is rare for most models ($\approx 0\%$), with Qwen the exception ($62.8\%$ / $33.2\%$ on BI-RADS / density). Low accuracy does not necessarily reflect simple ignorance, however, the confidence and reasoning signals carry additional information about VLM behavior.

\subsection{Answer--Reasoning inconsistency}
\label{sec:ar}

\begin{figure}
\centering
\includegraphics[width=1.0\linewidth]{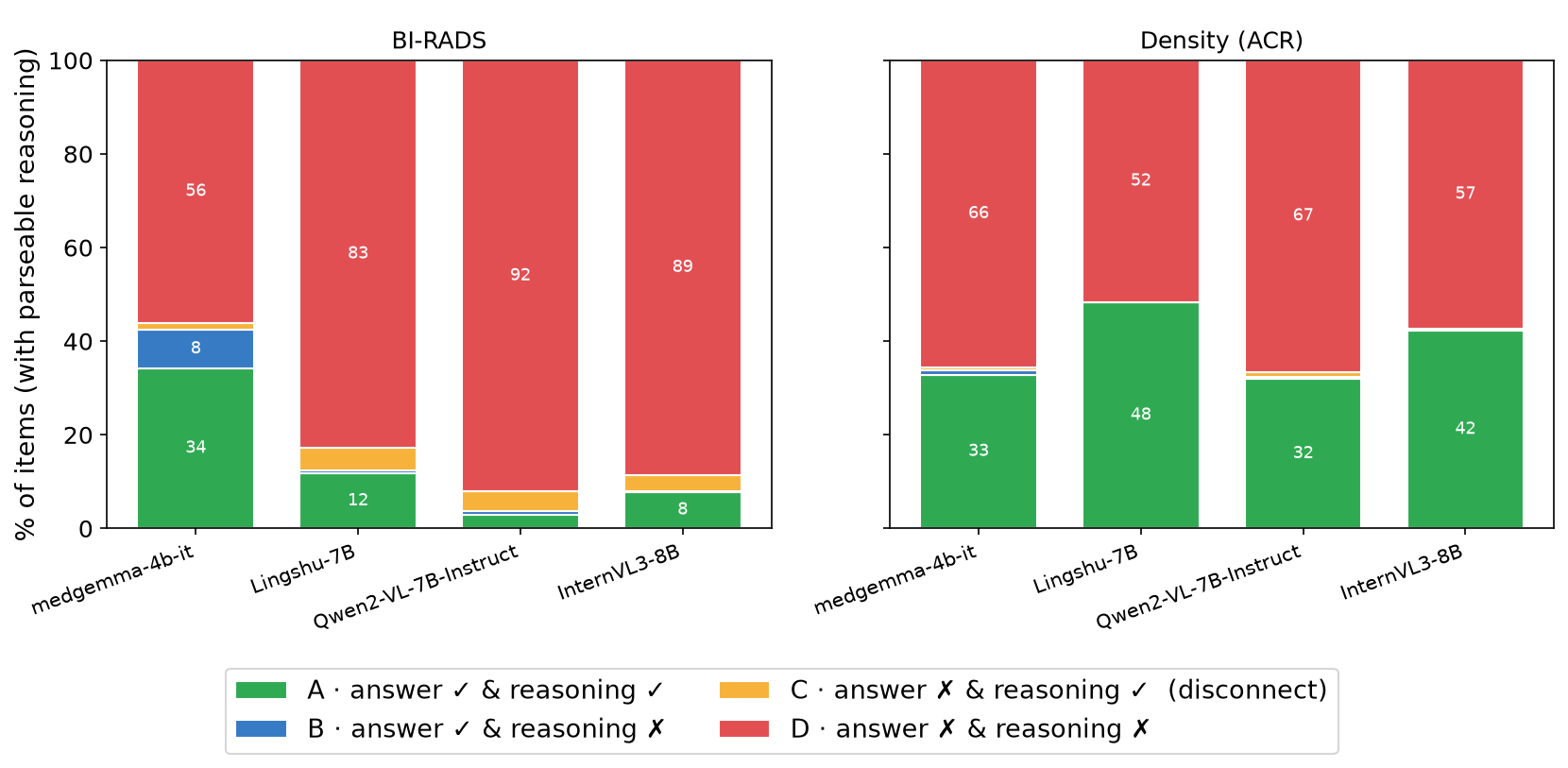}
\vspace{-18pt}
\caption{Answer $\times$ reasoning coherence on RSNA. Each item is classified by whether the explicit answer $\hat y = y$ and the reasoning answer $\hat y_e = y$: A (both correct), B (answer correct, reasoning wrong), C (answer wrong, reasoning correct; the \emph{disconnect}), D (both wrong).}
\label{fig:taxonomy}
\end{figure}

Elicited directly, a model's explicit answer and its written rationale are largely \emph{coherent}: the taxonomy (Figure~\ref{fig:taxonomy}) is dominated by the two agreement buckets, A (both correct) and D (both wrong), while the disconnect bucket~C (answer wrong but reasoning correct) is negligible ($<5\%$ for every model/task). Coherence, however, is not competence. The largest bucket on BI-RADS is
\emph{D}---the model's answer and its reasoning agree and are \emph{both wrong}---at $56\%$ (MedGemma), $83\%$ (Lingshu), $89\%$ (InternVL), and $92\%$ (Qwen), against a coherent-correct bucket A of only $3$--$34\%$. Density is less extreme (A $=32$--$48\%$, D $=52$--$67\%$) but shows the same pattern.

\begin{figure*}
\centering
\includegraphics[width=1.0\linewidth]{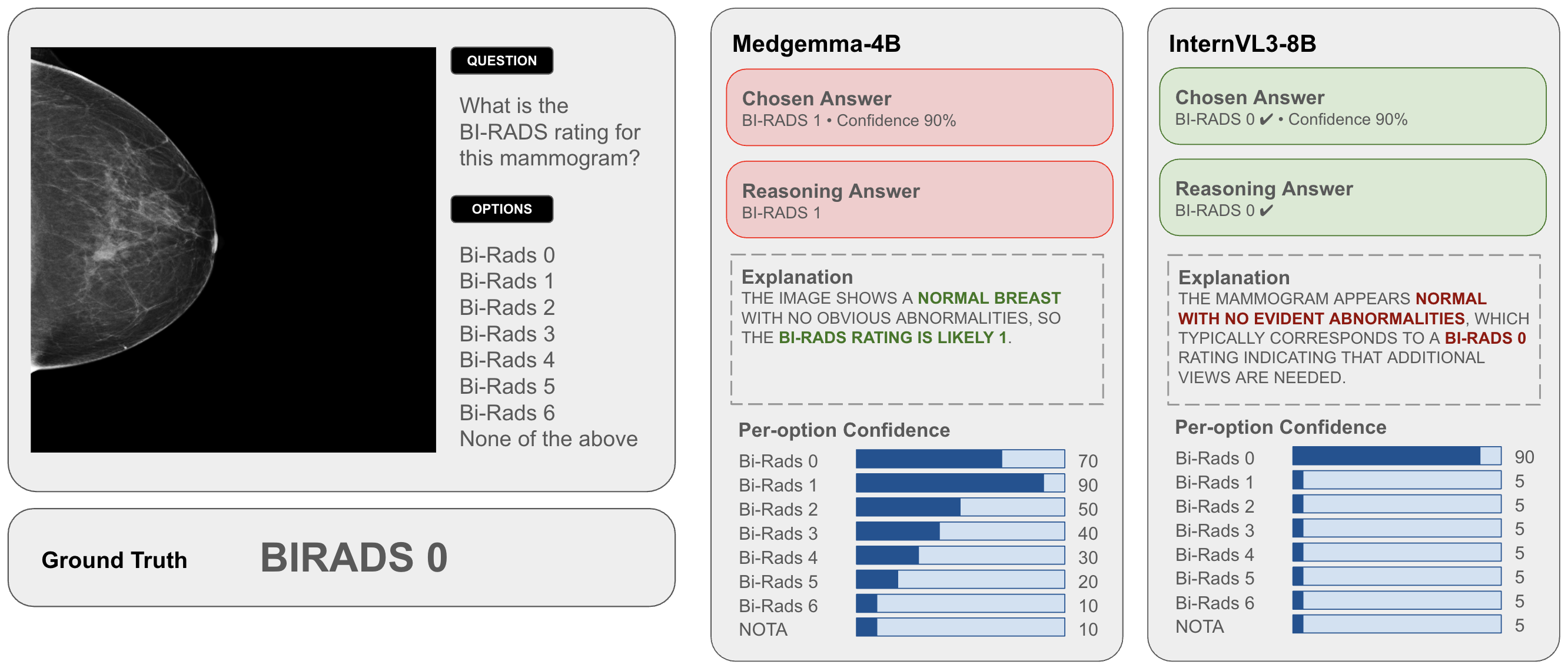}
\vspace{-8pt}
\caption{Two models on the same RSNA mammogram (ground truth BI-RADS~0).
Each model's stated answer agrees with the category implied by its explanation, yet the \emph{correct}
answer (InternVL, BI-RADS~0) rests on a self-contradictory rationale---it calls the exam ``normal''
(a BI-RADS~1 finding) while assigning BI-RADS~0---whereas the \emph{incorrect} answer (MedGemma,
BI-RADS~1) follows validly from the same normal reading.}
\label{fig:disconnect_case}
\end{figure*}

\textbf{Coherent does not mean valid.}  Within the agreement buckets, the findings a model describes often fail to warrant the category it assigns. Figure~\ref{fig:disconnect_case} shows this on a single screening exam (ground truth BI-RADS~0). InternVL assigns BI-RADS~0 and is therefore scored
\emph{correct}, yet its rationale describes the breast as normal with no abnormalities, a BI-RADS~1
reading, so the category it reports contradicts the finding it describes. MedGemma reads the exam the
same way but, reasoning validly from ``normal,'' assigns BI-RADS~1 and is scored \emph{wrong}. Accuracy
thus credits the model whose reasoning is self-contradictory and penalises the one whose reasoning is
sound: a correct label is no evidence that the model read the image correctly, and only inspecting the
rationale exposes the disconnect. This is not an isolated case. Across the agreement buckets, the findings a model describes routinely fail to justify the category it assigns. Additional results are in Appendix~\ref{app:ar-gap}.

\subsection{Confidence--evidence inconsistency}
\label{sec:conf}

\begin{figure*}
\centering
\includegraphics[width=1\linewidth]{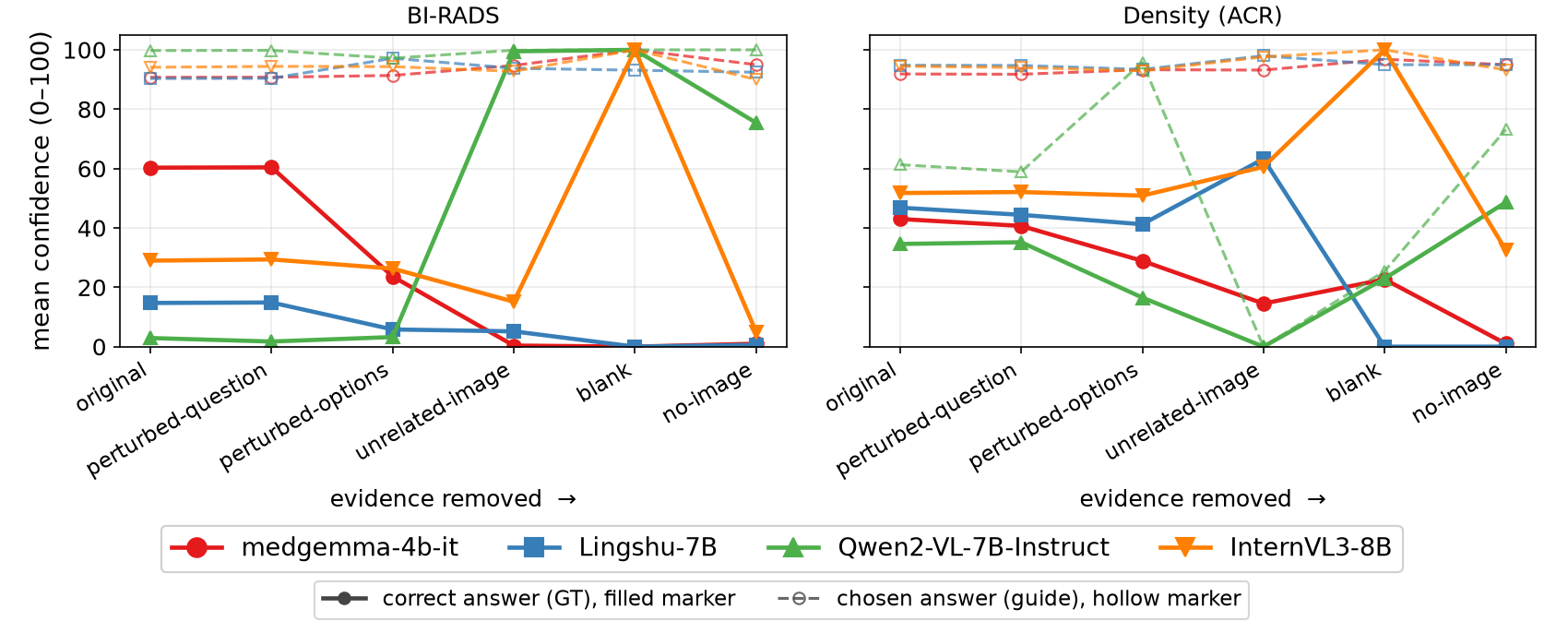}
\vspace{-8pt}
\caption{Mean confidence on the condition-correct answer, $c_y$ (solid) the original category on real-image conditions, NOTA on the visual perturbations vs.\ on the explicitly chosen answer, $c_{\hat y}$ (dashed), across the six RSNA conditions.}
\label{fig:gt_conf}
\end{figure*}

Figure~\ref{fig:gt_conf} reports the mean confidence each model assigns to the condition-correct answer, $c_y$ (solid), and to the option it explicitly chose, $c_{\hat y}$ (dashed). The correct answer is the original category on the real-image conditions and NOTA on the three visual perturbations. A
grounded, calibrated model should maintain a high $c_y$ across all conditions, confident in the category when the mammogram is present, and in NOTA once it is removed.

The models instead keep $fc_y$ low while $c_{\hat y}$ remains near $90$--$100$, indicating confidence that is consistent but misdirected. Two opposite failure modes emerge. MedGemma is confident in the correct category on the real image (BI-RADS $c_y=60.3$), yet this collapses to near zero under the visual
perturbations, where it keeps backing a category ($c_{\hat y}\approx95$) instead of abstaining. Qwen does the reverse: it defaults to NOTA almost regardless of the image, so $c_y$ is near zero on the real image ($2.9$) and high on the junk conditions ($75$--$100$), resulting correct only because it always abstains. Density follows the same split (MedGemma $43.0\rightarrow1.0$; Lingshu $46.8\rightarrow0.0$ without an image). In every case the models are confident, however, in none does that confidence track the evidence.

\begin{table}[t]
\centering
\caption{RSNA ``None of the above'' selection rate on the real image vs.\ its average over the
three visual perturbations (\textsc{unrelated}/\textsc{blank}/\textsc{no image}), where NOTA is the
correct response, with $\Delta=$ Visual $-$ Real.}
\label{tab:nota}
\scriptsize
\setlength{\tabcolsep}{4pt}
\begin{tabular}{lrrr}
\toprule
Model & Real \% & Visual \% & $\Delta$ (pp) \\
\midrule
\multicolumn{4}{l}{\textit{BI-RADS}} \\
MedGemma-4B   & 0.0  & 0.0  & $+0.0$  \\
Lingshu-7B    & 0.0  & 20.6 & $+20.6$ \\
Qwen2-VL-7B   & 62.8 & 91.7 & $+28.9$ \\
InternVL3-8B  & 13.7 & 91.0 & $+77.3$ \\
\midrule
\multicolumn{4}{l}{\textit{Density (ACR)}} \\
MedGemma-4B   & 0.0  & 12.7 & $+12.7$ \\
Lingshu-7B    & 0.0  & 20.9 & $+20.9$ \\
Qwen2-VL-7B   & 33.2 & 81.5 & $+48.3$ \\
InternVL3-8B  & 0.0  & 75.3 & $+75.3$ \\
\bottomrule
\end{tabular}
\end{table}

\subsection{NOTA selection}
\label{sec:nota}

Table~\ref{tab:nota} reports the explicit NOTA-selection rate on the real image against its average over the three visual perturbations (\textsc{unrelated} / \textsc{blank} / \textsc{no image}), where NOTA is the correct response, alongside $\Delta_{\mathrm{NOTA}}$. Two regimes emerge.
\emph{Evidence-sensitive} models raise their NOTA rate substantially under perturbation: InternVL from $14$ to $91$ ($+77$) on BI-RADS, and Qwen from $63$ to $92$ ($+29$). \emph{Evidence-insensitive} models barely respond: MedGemma holds at $0$ on BI-RADS ($\Delta=+0$).

The shift alone, however, overstates how well these models abstain. Lingshu's $+21$ raises its visual NOTA rate only to $21\%$: it still returns a categorical answer to roughly four of every five invalid images. MedGemma is worse, abstaining on $0\%$ of junk BI-RADS images and $13\%$ on density; it assigns a definite category to essentially every image that carries no valid evidence. Only InternVL and Qwen abstain on the majority of junk ($75$--$92\%$), and even they commit to a category on up to a quarter of the density perturbations. Additional results are in Appendix~\ref{app:nota}.

\subsection{Perturbation sensitivity}
\label{sec:pert}

\begin{figure}[t]
\centering
\includegraphics[width=1.0\linewidth]{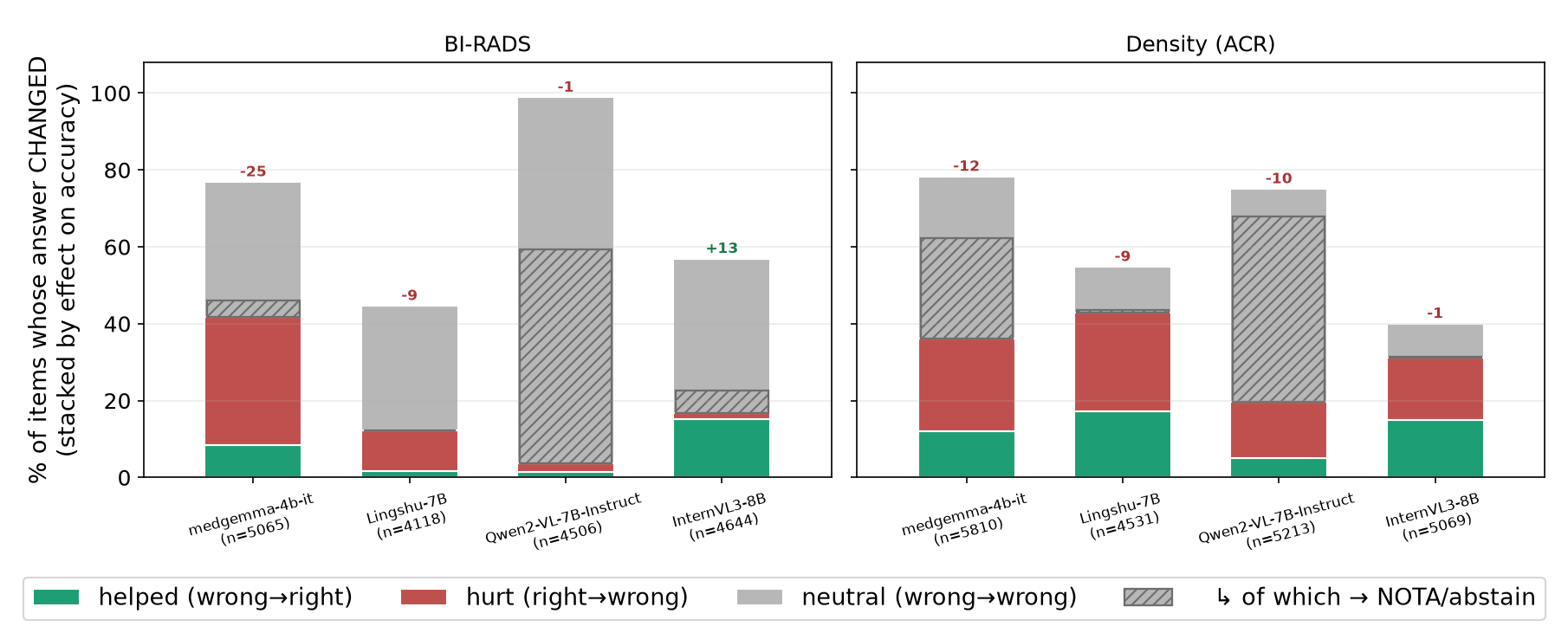}
\vspace{-18pt}
\caption{Perturbed-Options condition effect on RSNA. Bars = share of items whose stated answer changes when the options are reordered, split by effect on accuracy.}
\label{fig:pert_opt}
\end{figure}

Figure~\ref{fig:pert_opt} reports what happens when the answer options are shuffled, measured as the share of items whose stated answer changes, split by whether the change helps, hurts, or leaves accuracy unchanged. An answer read from the mammogram would be \emph{invariant} to the order in which the choices are listed; instead we see flip rates of
$40$--$99\%$ ($77\%$ for MedGemma and $57\%$ for InternVL on BI-RADS, $55$--$78\%$ on density). That a single reordering overturns most answers indicates that the choice is anchored to the surface arrangement of the options rather than to evidence in the image. The \emph{direction} of the flips makes the same point: they are overwhelmingly \emph{neutral} (wrong$\rightarrow$wrong), so reordering reshuffles an already-ungrounded guess rather than correcting it, and where accuracy does move it predominantly \emph{hurts} (net $-24.8$ for MedGemma to $+13.4$ for InternVL on BI-RADS). Qwen is the most volatile, flipping on $99\%$ of BI-RADS and $75\%$ of density items almost entirely by retreating to a newly-introduced NOTA/filler option ($56\%$ and $48\%$ of all items): its answer tracks whichever safe option is present, not the image. The asymmetries are significant under McNemar's test (Appendix~\ref{sec:appendix_pert}). Rewording the question, by contrast, leaves most answers intact (Appendix~\ref{sec:appendix_pert_q}), so the fragility is specific to option structure, not phrasing.
\section{Discussion}

Our results show that a model's answer, confidence, and explanation are largely decoupled from the
imaging evidence, and that coherence among these signals does not imply sound reasoning. This is a different problem from the one most medical VLM evaluation is built to catch. Accuracy-based benchmarks assume a wrong answer reflects a gap in knowledge or perception, but the gap here can instead sit within the reasoning itself, or between the stated answer and the explanation. We cannot distinguish whether the explanation reflects the model's actual decision process or is a post-hoc rationalization generated somewhat independently of it. Prior work on chain-of-thought faithfulness suggests the latter is common \citep{Turpin2023,Lanham2023}, and recent work on visual grounding of VLM rationales finds explanations are often produced after only a superficial inspection of the image, with individual steps unsupported by the evidence they cite \citep{yi2025corgi}. Either way, the explanation is a signal a clinician would plausibly act on, so its disagreement with the stated answer, or its invalidity, is a practical problem for trust and auditability regardless of which signal is closer to the model's true internal state. This same ambiguity extends to abstention: a high refusal rate under evidence removal can reflect genuine grounding or a fixed refusal policy, and only comparing it against the real-image rate tells the two apart.

\subsection{Limitations and Future Work.}
Our evaluation relies solely on MammoVQA's templated multiple-choice formulation, so some option-order and wording sensitivity may reflect the task framing. Extending the protocol to free-response or report-generation formats would test its generality. We also evaluate open, general-purpose and medically-adapted VLMs because mammography-specialized systems have not released their weights; whether domain-specific fine-tuning changes the coherence failures remains an open question.

A separate direction concerns the confidence-evidence inversion we observe. This may relate to mirage reasoning, where multimodal models produce confident clinical descriptions without image input \citep{asadi2026mirage}. Testing explicit versus implicit image-removal prompts could help distinguish these phenomena.

Our perturbations omit clinically relevant changes such as swapped laterality or view, and same-patient prior images. Post-hoc verification methods could test whether reasoning-correct, answer-wrong items reflect visual grounding \citep{yi2025corgi}. Our reasoning matcher was validated on a modest sample, and parse rates vary across models. Future work could compare verbalized with logit-based confidence, since independent option scores do not necessarily form a distribution the way normalized probabilities do.

\bibliography{sn-bibliography}

\begin{appendices}
\section{Model Configurations}
\label{app:model-settings}

Table~\ref{tab:model_configs} lists the checkpoint and inference settings for each evaluated model.

\begin{table}[tbp]
\centering
\scriptsize
\caption{Model configurations. All models use the default (latest) Hugging Face checkpoint revision, bfloat16 precision (no quantization), greedy decoding (temperature $0$), and a 384-token generation budget.}
\label{tab:model_configs}
\begin{tabular}{lllc}
\toprule
Model & Hugging Face checkpoint & Domain & Params \\
\midrule
MedGemma-4B  & \texttt{google/medgemma-4b-it}           & Medical & 4B \\
Lingshu-7B   & \texttt{lingshu-medical-mllm/Lingshu-7B} & Medical & 7B \\
Qwen2-VL-7B  & \texttt{Qwen/Qwen2-VL-7B-Instruct}       & General & 7B \\
InternVL3-8B & \texttt{OpenGVLab/InternVL3-8B}          & General & 8B \\
\bottomrule
\end{tabular}
\end{table}

\section{Dataset Descriptions}
\label{app:data}

Table \ref{tab:data-breakdown} reports the item count and majority-class prior for each cohort-task pair evaluated in this work. The majority-class prior is computed as the accuracy of a trivial classifier that always predicts the single most frequent label in that cohort-task's label distribution over the benchmark items, and serves as the reference point for VLM performance. EMBED is evaluated only on the BI-RADS assessment task. Full per-category label distributions underlying each majority-class prior, rather than just the majority share, are available on request but omitted here for space.

\begin{table}[h]
\centering
\caption{Cohort and task composition. Majority-class prior is the
accuracy of always predicting the most frequent label.}
\label{tab:data-breakdown}
\scriptsize
\begin{tabular}{llrl}
\toprule
Cohort & Task & $N$ & Majority-class prior \\
\midrule
RSNA  & BI-RADS        & 5{,}106 & 59.4\% \\
RSNA  & Density  &  5{,}821 & 42.6\% \\
EMBED & BI-RADS        & 14,303      & 62.1\% \\
\bottomrule
\end{tabular}
\end{table}

\section{Additional RSNA Results}
\label{app:rsna}

\subsection{Answer vs.\ Reasoning}
\label{app:ar-gap}

\begin{figure*}
\label{app:parser_examples}
\centering
\includegraphics[width=0.65\linewidth]{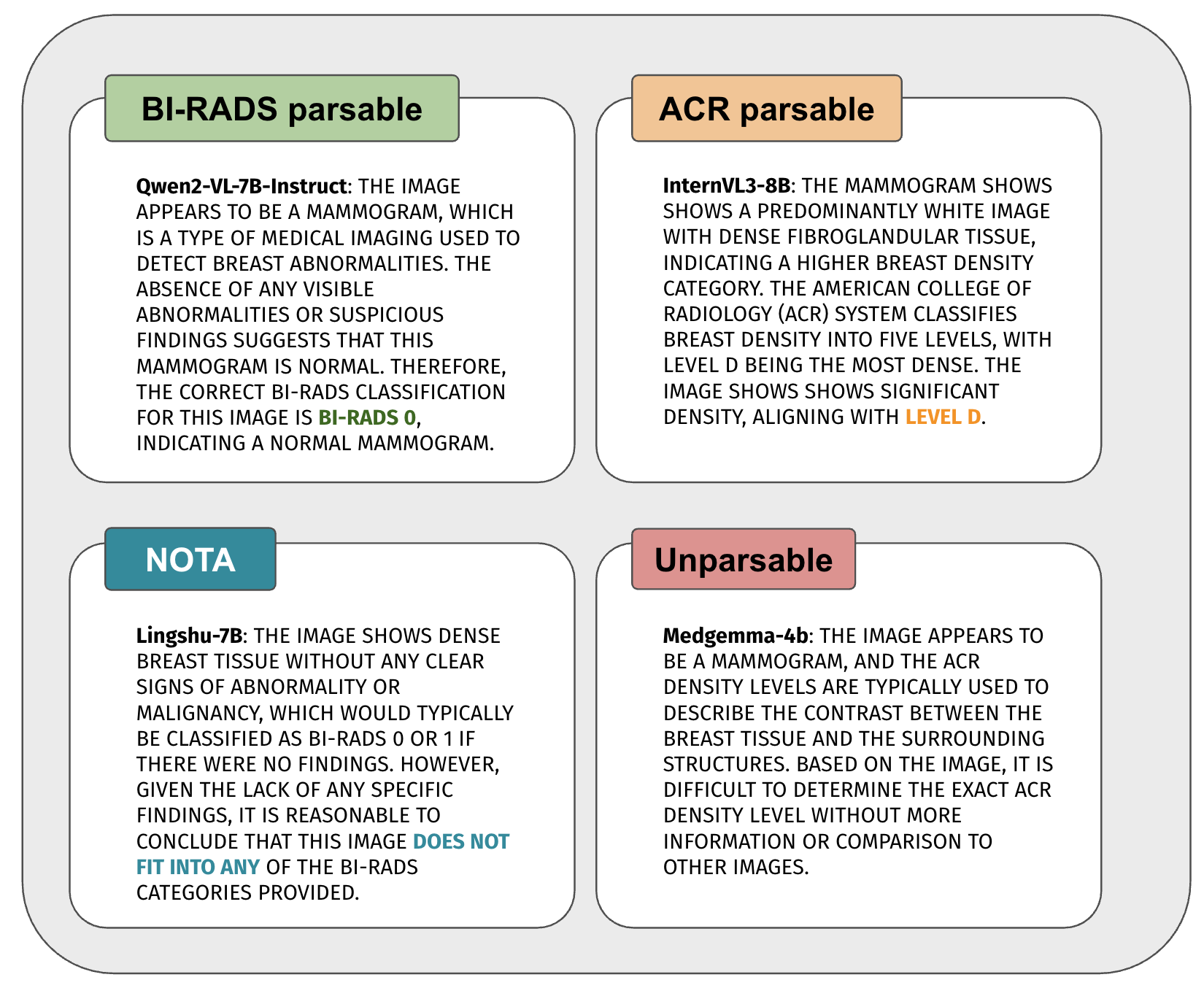}
    \caption{Example explanations for the four reasoning-parser outcomes: BI-RADS label, ACR-density label, ``None of the above," and unparsable. Highlighted text is the phrase the parser matches.}
    \label{fig:parser_example}
\end{figure*}

Table~\ref{tab:ar_gap} reports explicit-answer and reasoning accuracy per model and task. Because both accuracies here are computed only on the parseable subset, they are not directly comparable to the unconditioned accuracies reported in Table~\ref{tab:baseline}. Figure~\ref{fig:parser_example} gives representative explanations for the parser's four outcomes. We use a deterministic, released rule-based matcher as the primary reasoning evaluator: it is reproducible, matches explanations against the standardized BI-RADS/ACR lexicon, and scales to $\sim$$950$k generations, whereas an LLM judge could itself exhibit the incoherence this paper studies. On $120$ hand-labeled explanations the matcher matches a blinded annotator at $\kappa=0.78$ ($80.8\%$), comparably to an LLM judge (\texttt{gpt-oss-20b}) blinded to the ground truth and the matcher (Table~\ref{tab:extract_human}). At scale, the judge matches the matcher on $\approx$$70.6\%$ of $\sim$$8{,}000$ explanations ($58$--$94\%$ per cell; Table~\ref{tab:extract_scale}). Disagreements fall in both directions, mostly on rationales that describe a finding without naming a category, so the matcher's rates are a conservative lower bound.

\begin{table}[tbp]
\centering
\scriptsize
\caption{Explicit answer vs.\ reasoning accuracy on the parsable fraction of \textbf{RSNA} results.
Both accuracies are computed only on items whose explanation yields a readable category, so the
comparison within each row is paired. Parse rate varies across model$\times$task cells, from $66\%$
(Qwen2-VL-7B, BI-RADS) to $\sim$$100\%$ (MedGemma-4B, BI-RADS); low-parse cells rest on a
correspondingly smaller subset. The gap $\Delta=(\text{reasoning}-\text{answer})$ is small and mixed in
sign ($-6.7$ to $+4.3$): the free-text rationale is no more accurate than the explicitly stated
answer.}
\label{tab:ar_gap}
\begin{tabular}{llrrrr}
\toprule
Model & Task & Parsed \% & Answer acc.\ \% & Reasoning acc.\ \% & $\Delta$ (pp) \\
\midrule
MedGemma-4B  & BI-RADS & 99.8 & 42.4 & 35.8 & $-6.7$ \\
             & Density & 94.1 & 33.8 & 33.5 & $-0.3$ \\
\addlinespace
Lingshu-7B   & BI-RADS & 97.6 & 12.3 & 16.6 & $+4.3$ \\
             & Density & 92.2 & 48.4 & 48.3 & $+0.0$ \\
\addlinespace
Qwen2-VL-7B  & BI-RADS & 66.3 & 3.7  & 7.1  & $+3.4$ \\
             & Density & 75.9 & 32.5 & 32.9 & $+0.5$ \\
\addlinespace
InternVL3-8B & BI-RADS & 96.4 & 7.9  & 11.4 & $+3.5$ \\
             & Density & 97.5 & 42.5 & 42.5 & $+0.0$ \\
\bottomrule
\end{tabular}
\end{table}

\begin{table}[tbp]
\centering
\scriptsize
\caption{Reasoning-extraction agreement on $120$ hand-labeled RSNA explanations.}
\label{tab:extract_human}
\begin{tabular}{lrr}
\toprule
Comparison & Agreement \% & $\kappa$ \\
\midrule
Matcher vs.\ human & 80.8 & 0.777 \\
Judge vs.\ human   & 82.5 & 0.798 \\
Matcher vs.\ judge & 78.3 & 0.753 \\
\bottomrule
\end{tabular}
\end{table}

\begin{table}[tbp]
\centering
\scriptsize
\caption{Matcher vs.\ LLM-judge agreement at scale ($1{,}000$ explanations per task, judge blinded).
Overall $\approx$$70.6\%$.}
\label{tab:extract_scale}
\begin{tabular}{llr}
\toprule
Model & Task & Agreement \% \\
\midrule
MedGemma-4B  & BI-RADS & 93.5 \\
             & Density & 74.5 \\
Lingshu-7B   & BI-RADS & 58.0 \\
             & Density & 66.7 \\
Qwen2-VL-7B  & BI-RADS & 84.9 \\
             & Density & 59.1 \\
InternVL3-8B & BI-RADS & 60.9 \\
             & Density & 67.0 \\
\bottomrule
\end{tabular}
\end{table}

\subsection{NOTA selection and confidence across all conditions}
\label{app:nota}

\begin{figure}[t]
\centering
\includegraphics[width=1.0\linewidth]{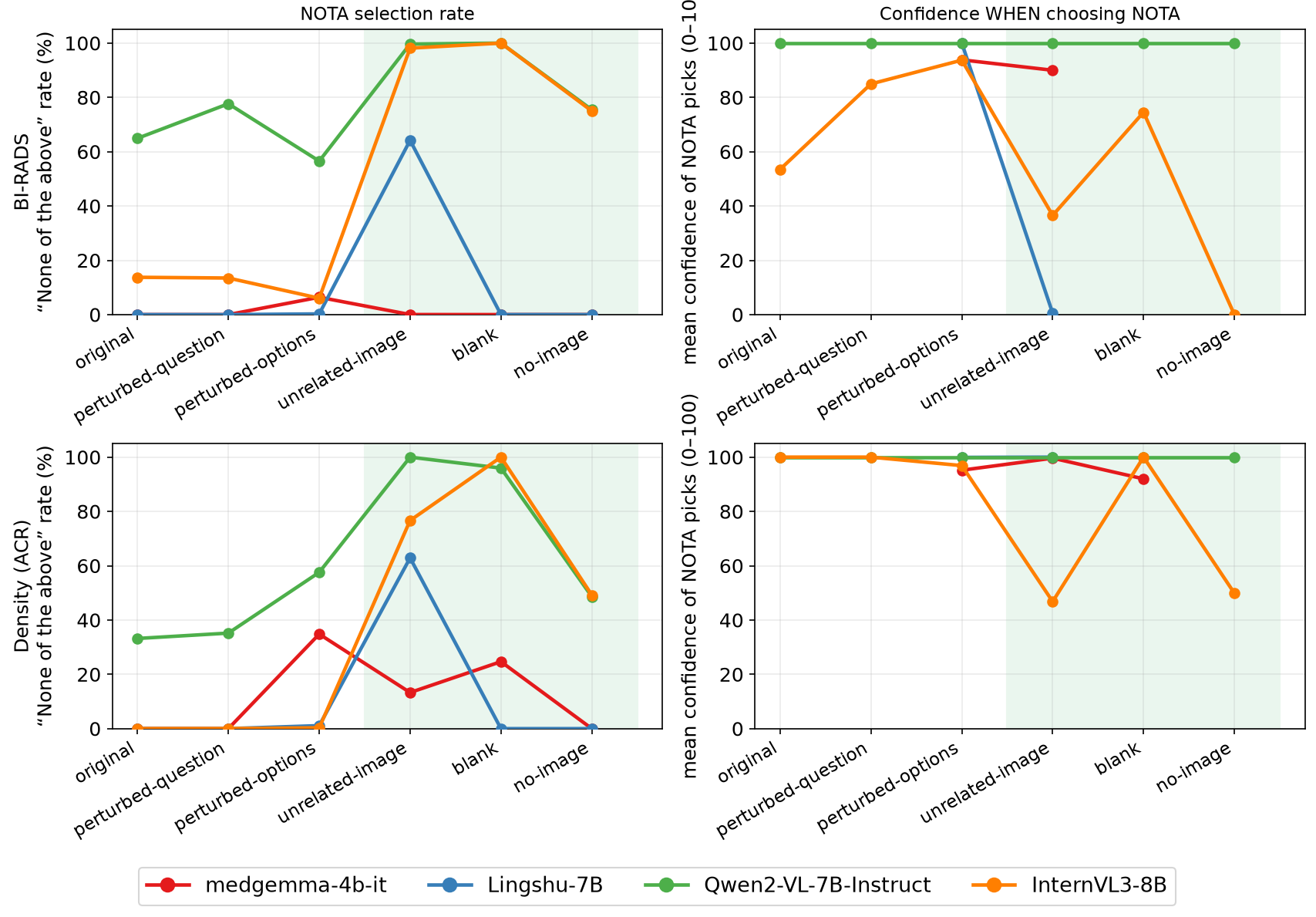}
\caption{NOTA-selection rate (left) and mean confidence when NOTA is chosen
(right), for BI-RADS (top) and Density/ACR (bottom), across all six
conditions. Shaded band marks the three conditions where NOTA is the
correct response.}
\label{fig:nota_confidence}
\end{figure}

Figure~\ref{fig:nota_confidence} reports the NOTA-selection rate and the confidence placed on NOTA across all six conditions. Only InternVL and Qwen raise their NOTA rate substantially under the visual perturbations ($75$--$100\%$ on BI-RADS); Lingshu abstains only on the unrelated photo, not on blank or no-image; and MedGemma barely abstains at all ($0\%$ on BI-RADS junk, $\le25\%$ on density). Most models thus do not treat the loss of a valid image as grounds to abstain. When NOTA is chosen its confidence is near $100$ for Qwen in every condition, but InternVL and Lingshu sometimes abstain at
very low confidence (InternVL $\approx0$ on no-image, Lingshu $\approx0.5$ on the unrelated photo), a fallback rather than a deliberate call.

\subsection{Question Rewording}
\label{sec:appendix_pert_q}
\begin{figure}[!ht]
\centering
\includegraphics[width=1.0\linewidth]{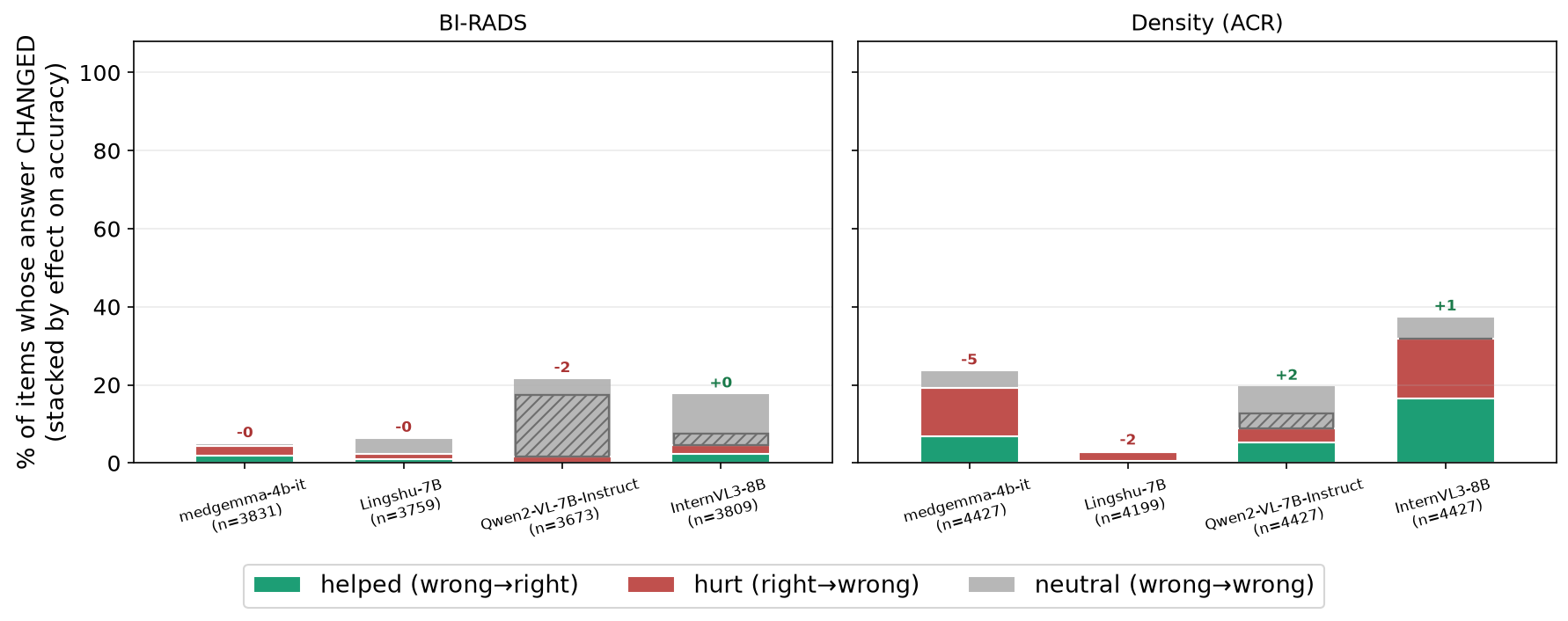}
\caption{Effect of rewording the question, on items whose wording actually changed. Each bar is the
share of items whose answer changed, split by effect on accuracy. The cross-hatched portions are
the share that changed to a NOTA. Numbers above the bars give the net accuracy change (percentage points).}
\label{fig:pert_q}
\end{figure}

Rewording the question (restricted to items whose wording actually changed) barely moves accuracy: the
net change is small for every model, from $-5.4$ points (MedGemma, density) to $+1.8$ (Qwen, density).
Answers still change more often than this net effect suggests, though far less than under option
shuffling: rewording changes $5$--$22\%$ of BI-RADS answers and $3$--$38\%$ of density answers
(Figure~\ref{fig:pert_q}). Most of these
changes are from one wrong answer to another (for Qwen, much of the BI-RADS change ($15.8$ of
$21.8\%$) is a switch to NOTA) so rewording alters answers without making them more correct. InternVL
is the most sensitive to rewording (a $37.6\%$ change rate on density), yet even there accuracy rises
only $+1.3$ points. Because rewording preserves the meaning of the question, a model that read the image should give the same answer. Answers changing at all, shows the fragility lies in the surface form of the task.

\subsection{Perturbation Sensitivity Details}
\label{sec:appendix_pert}

\begin{table}[tbp]
\centering
\scriptsize
\setlength{\tabcolsep}{4pt}
\caption{McNemar test results for the two elicitation perturbations (reworded question, shuffled
options), per model and task, on the explicitly stated answer. Density refers to ACR breast density.
C$\rightarrow$W and W$\rightarrow$C are item counts; Net $\Delta$ is the percentage-point change in
accuracy. *: $p<0.05$.}
\label{tab:mcnemar}
\begin{tabular}{lllrrrrr}
\toprule
Model & Task & Pert. & Orig.\% & Pert.\% & C$\to$W & W$\to$C & Net $\Delta$ \\
\midrule
MedGemma-4B  & BI-RADS & Question & 46.0 & 45.6 & 88   & 75  & $-0.3$   \\
             & BI-RADS & Options  & 42.5 & 17.7 & 1685 & 431 & $-24.8$* \\
             & Density & Question & 31.5 & 26.0 & 544  & 303 & $-5.4$*  \\
             & Density & Options  & 33.9 & 21.7 & 1406 & 697 & $-12.2$* \\
\addlinespace
Lingshu-7B   & BI-RADS & Question & 10.5 & 10.0 & 52   & 36  & $-0.4$   \\
             & BI-RADS & Options  & 14.3 & 5.5  & 433  & 68  & $-8.9$*  \\
             & Density & Question & 50.2 & 48.6 & 92   & 24  & $-1.6$*  \\
             & Density & Options  & 48.5 & 39.8 & 1169 & 777 & $-8.7$*  \\
\addlinespace
Qwen2-VL-7B  & BI-RADS & Question & 3.0  & 1.4  & 60   & 1   & $-1.6$*  \\
             & BI-RADS & Options  & 2.4  & 1.6  & 103  & 68  & $-0.8$*  \\
             & Density & Question & 32.4 & 34.2 & 156  & 235 & $+1.8$*  \\
             & Density & Options  & 25.4 & 15.9 & 760  & 263 & $-9.5$*  \\
\addlinespace
InternVL3-8B & BI-RADS & Question & 8.3  & 8.3  & 88   & 89  & $+0.0$   \\
             & BI-RADS & Options  & 7.7  & 21.0 & 81   & 701 & $+13.4$* \\
             & Density & Question & 44.1 & 45.4 & 675  & 731 & $+1.3$   \\
             & Density & Options  & 41.8 & 40.8 & 813  & 763 & $-1.0$   \\
\bottomrule
\end{tabular}
\end{table}

For each model$\times$task$\times$perturbation cell, we test whether a perturbation induces an \emph{asymmetric} change in correctness using McNemar's exact test on the paired outcomes (correct/incorrect before vs.\ after perturbation). Let $n_{C\to W}$ and $n_{W\to C}$ denote the number of items that flip from correct to wrong and from wrong to correct, respectively, under the perturbation. Items whose correctness is unchanged (correct$\to$correct, wrong$\to$wrong) are
uninformative for this test and excluded. Under the null hypothesis that a perturbation is equally likely to help or hurt an initially-discordant item, $n_{C\to W} \sim
\mathrm{Binomial}(n_{C\to W}+n_{W\to C},\,0.5)$; we report the exact two-sided binomial $p$-value and
flag $p<0.05$ with an asterisk in Table~\ref{tab:mcnemar}. A significant result indicates the perturbation reliably pushes answers in one direction (typically correct$\to$wrong), even when the net accuracy change is small; both flip directions can be individually large and nearly cancel (Qwen on density: net $+1.8$ yet significant). Option shuffling produces a significant asymmetry in seven of the eight cells, often with a large net drop (down to $-24.8$ for MedGemma on BI-RADS).

\section{Additional EMBED Results}
\label{app:embed}

We repeat the four main-text analyses on EMBED BI-RADS (up to $N{=}14{,}303$ items;
majority-class prior $62.1\%$, ``BI-RADS~1''). EMBED is
\emph{exam-level}: each instance is a full mammography exam of \emph{1--15} views rather than a single image, and the model sees all views at once. The visual perturbations preserve
this structure---each exam's views are replaced by the same number of blank images, or by
the same number of distinct unrelated photos---so an exam of $n$ real views becomes $n$
blank (or $n$ unrelated) images. The trends broadly follow RSNA and are often more extreme, driven by EMBED's stronger class imbalance and exam-level setting.

\subsection{Answer--reasoning inconsistency.}

\begin{figure}[t]
\centering
\includegraphics[width=1.0\linewidth]{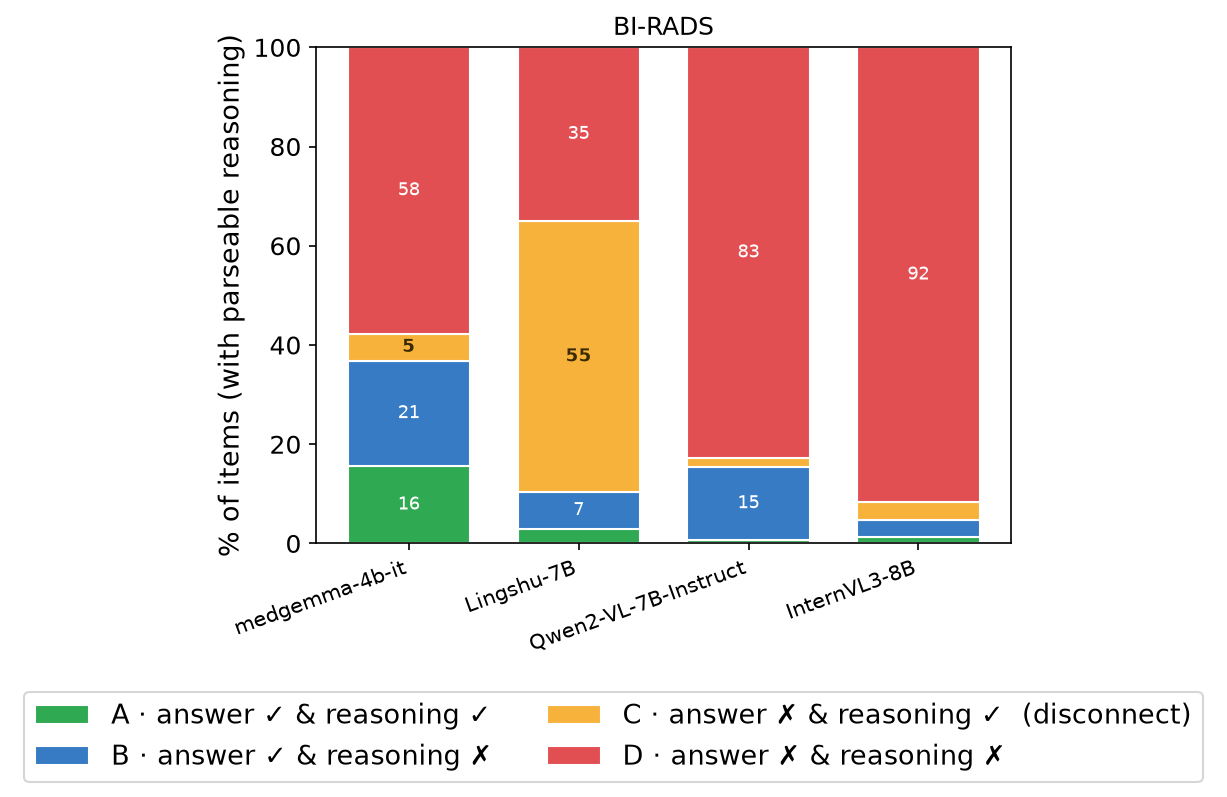}
\caption{Answer\,$\times$\,reasoning coherence (EMBED, BI-RADS). Orange = disconnect (answer wrong, reasoning correct).}
\label{fig:embed-taxonomy}
\end{figure}

Answer accuracy stays far below the class prior ($4.8$--$36.8\%$ vs.\ $62.1\%$;
Table~\ref{tab:embed-outreason}). For MedGemma, Qwen and InternVL the answer and its reasoning are
largely coherent and jointly wrong: bucket~D dominates ($58$--$92\%$) and the disconnect bucket~C is
small ($\le 5.4\%$; Fig.~\ref{fig:embed-taxonomy}). Lingshu is the exception: its explanation names the
correct category far more often than its stated answer ($57.6\%$ vs.\ $10.3\%$), a large disconnect
($C{=}54.6\%$), resulting in an output collapse in which it commits to a fixed wrong category regardless of a
correct rationale.

\begin{table}[tbp]
\centering
\scriptsize
\caption{Explicit answer vs.\ reasoning accuracy on the parsable fraction of \textbf{EMBED} BI-RADS,
both computed on items with parseable reasoning (paired). Parsed \% is relative to the $\sim$$14.3$k
item set; $\Delta=$ reasoning $-$ answer.}
\label{tab:embed-outreason}
\begin{tabular}{lrrrr}
\toprule
Model & Parsed \% & Answer acc.\ \% & Reasoning acc.\ \% & $\Delta$ (pp) \\
\midrule
MedGemma-4B   & 96.3 & 36.8 & 20.9 & $-15.9$ \\
Lingshu-7B    & 99.8 & 10.3 & 57.6 & $+47.3$ \\
Qwen2-VL-7B   & 91.4 & 15.4 & 2.3  & $-13.1$ \\
InternVL3-8B  & 82.4 & 4.8  & 4.7  & $-0.1$  \\
\bottomrule
\end{tabular}
\end{table}

\subsection{Confidence--evidence inconsistency.}
\begin{figure}[t]
\centering
\includegraphics[width=1.0\linewidth]{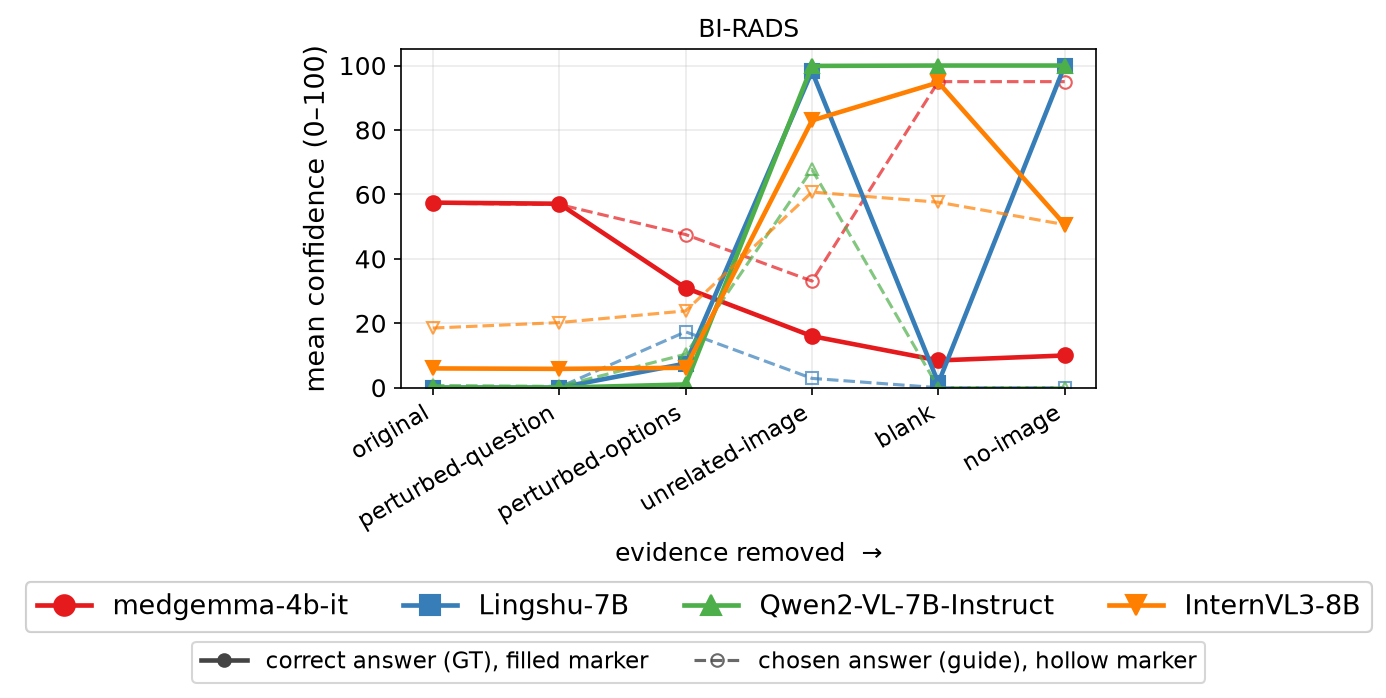}
\caption{Mean confidence on the condition-correct answer $c_y$ (solid; original category on real images, NOTA on the visual perturbations) vs.\ the explicitly chosen answer $c_{\hat y}$ (dashed), across the six EMBED (BI-RADS) conditions.}
\label{fig:embed-confreason}
\end{figure}

MedGemma's chosen-answer confidence \emph{rises} as the image is removed ($57$ on the real image $\to 95$ on a blank or absent image) while its confidence on the correct label falls to $\le 10$; InternVL likewise grows more confident on junk ($18.5\to57.6$). Lingshu and Qwen place near-zero confidence on both the chosen and the correct option at the real image. As in the main text, the declared confidence does not track whether an answerable image is present (Fig.~\ref{fig:embed-confreason}).

\subsection{NOTA selection.}
\begin{figure}[t]
\centering
\includegraphics[width=0.8\linewidth]{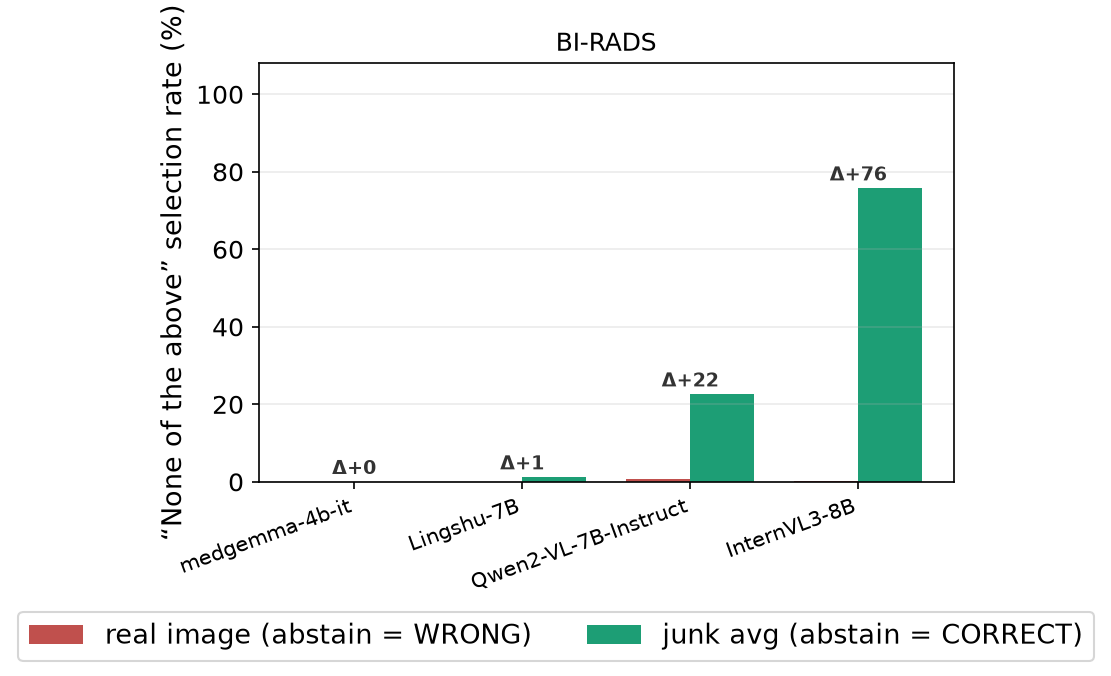}
\caption{NOTA-selection rate on the real image vs.\ junk images (EMBED, BI-RADS). Evidence-driven NOTA should rise under junk; most models do not.}
\label{fig:embed-nota}
\end{figure}

At the real image the models rarely abstain ($\le 0.7\%$). Under the visual perturbations only InternVL
raises its NOTA rate substantially ($0.1\to75.8\%$ on average, reaching $100\%$ on no-image); Qwen abstains only on the unrelated photo ($67.9\%$, versus $0\%$ on blank and no-image), while MedGemma and Lingshu almost never abstain even on junk ($\le 3.7\%$). As on RSNA, abstention does not reliably track the loss of a valid image (Fig.~\ref{fig:embed-nota}).

\subsection{Perturbation sensitivity.}
\begin{figure}
\centering
\includegraphics[width=1.0\linewidth]{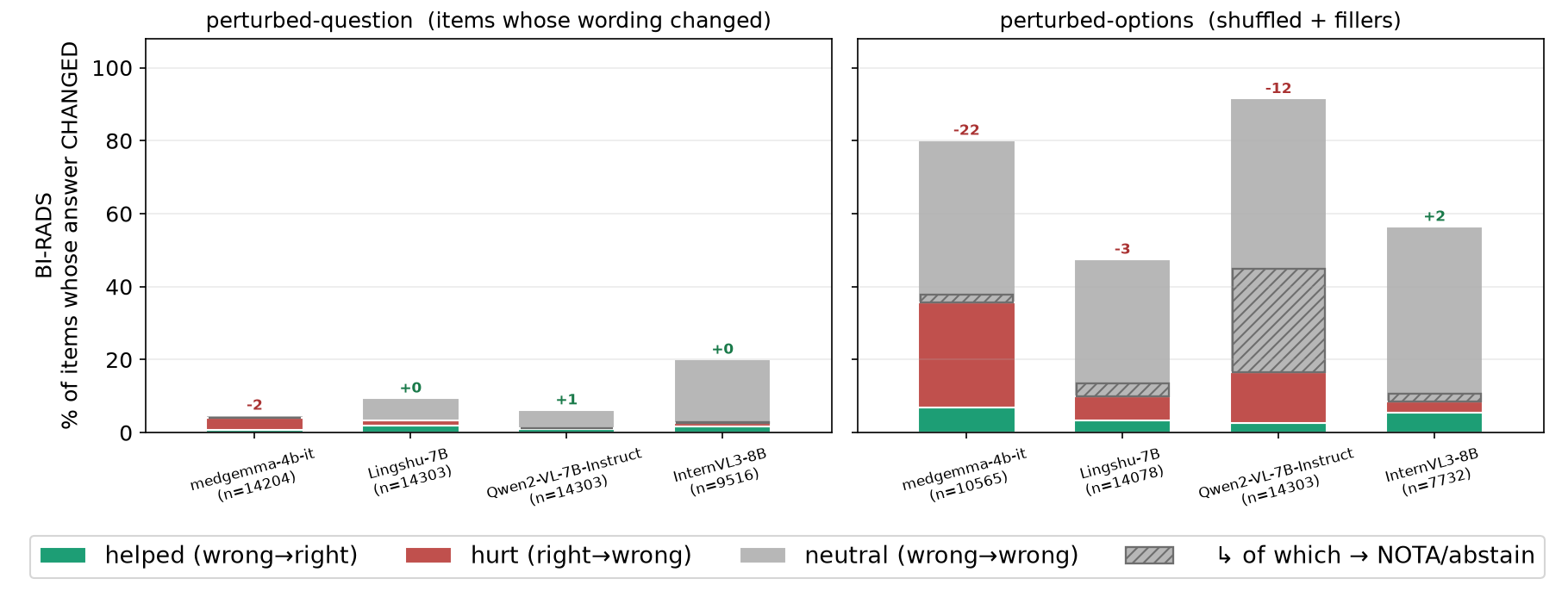}
\caption{Perturbation sensitivity (EMBED, BI-RADS): question wording (left) vs.\ option shuffling + fillers (right). Option shuffling flips most answers without raising accuracy.}
\label{fig:embed-pertflips}
\end{figure}

Rewording the question changes few answers ($5$--$20\%$, highest for InternVL) with little net effect
($|\Delta|\le 2.3$~pp). Shuffling the options and adding fillers changes $48$--$92\%$ of answers, predominantly from one wrong answer to another, often with a large net accuracy drop (MedGemma $-21.9$, Qwen $-11.6$~pp; Fig.~\ref{fig:embed-pertflips}), the same option-order fragility seen on RSNA (Section~\ref{sec:pert}).

\end{appendices}

\end{document}